\documentclass[letterpaper, 10 pt, conference]{ieeeconf}  
\usepackage{cite}
\usepackage{graphicx}
\usepackage{booktabs} 
\usepackage{bm}
\usepackage{amsmath}

\IEEEoverridecommandlockouts                              

\title{\LARGE \bf
Under-actuated Robotic Gripper with Multiple Grasping Modes Inspired by Human Finger
}

\author{Jihao Li$^{1}$$^,$$^{2}$, Tingbo Liao$^{3}$, Hassen Nigatu$^{1}$$^,$$^{2}$, Haotian Guo$^{1}$$^,$$^{2}$, Guodong Lu$^{2}$, and  Huixu Dong$^{1}$$^,$$^{2}$$^,$$^*$
\thanks{$^{1}$Grasp Lab, School of Mechanical Engineering, Zhejiang University, Hangzhou 310027, China.}%
\thanks{$^{2}$Robotics Institute of Zhejiang University, Hangzhou 310027, China.}%
\thanks{$^{3}$College of Design and Engineering National University of Singapore, 117575, Singapore.
}%
\thanks{Email: {\tt\small\{huixudong, lijihao\}@zju.edu.cn}}
\thanks{*Corresponding author}%
}

\begin{document}

\maketitle
\thispagestyle{empty}
\pagestyle{empty}

\begin{abstract}

Under-actuated robot grippers as a pervasive tool of robots have become a considerable research focus. Despite their simplicity of mechanical design and control strategy, they suffer from poor versatility and weak adaptability, making widespread applications limited. To better relieve relevant research gaps, we present a novel 3-finger linkage-based gripper that realizes retractable and reconfigurable multi-mode grasps driven by a single motor. Firstly, inspired by the changes occurred in the contact surface with a human finger moving, we artfully design a slider-slide rail mechanism as the phalanx to achieve retraction of each finger, allowing for better performance in the enveloping grasping mode. Secondly, a reconfigurable structure is constructed to broaden the grasping range of objects’ dimensions for the proposed gripper. By adjusting the configuration and gesture of each finger, the gripper can achieve five grasping modes. Thirdly, the proposed gripper is just actuated by a single motor, yet it can be capable of grasping and reconfiguring simultaneously. Finally, various experiments on grasps of slender, thin, and large-volume objects are implemented to evaluate the performance of the proposed gripper in practical scenarios, which demonstrates the excellent grasping capabilities of the gripper.

\end{abstract}
\section{INTRODUCTION}

As grasping is a crucial ability of robots, numerous robot grippers as grasping executive tools have been developed to be applied in various fields, such as logistics, warehousing scenarios, industrial settings, as well as human-computer interaction \cite{dong1,dong2,dong3}. Among various grippers, the full-actuated grippers have improved dexterity, owing to the independent motion of each joint \cite{full_actuated}. However, it brings about issues in the high cost and complexity of the mechanism and control. Therefore, under-actuated grippers have attracted substantial attention, being advantageous in simplifying the mechanical design, control strategy and enhancing grasping adaptability as well as reducing the cost \cite{under_actuated}.

 \begin{figure}[tp]
    \centering
    \vspace{0.15cm}
    \setlength{\abovecaptionskip}{-0.1cm}
    \includegraphics[width=8.4cm]{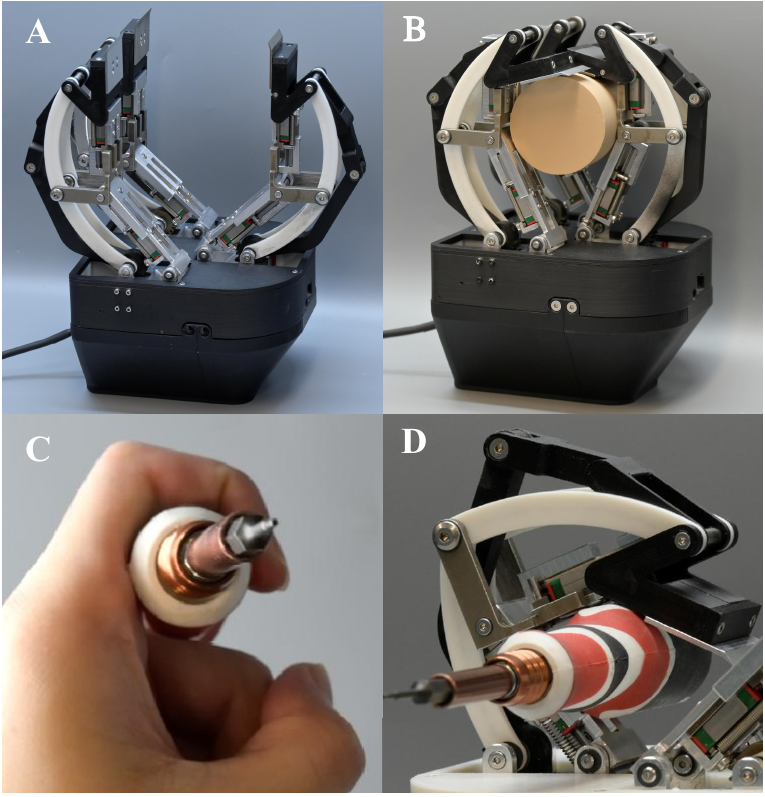}
    \caption{The prototype of presented gripper (A, B), grasp slender and thin objects through the human finger (C) and the presented gripper (D).}
    \label{figurelabel}
    \vspace{-0.8cm}
\end{figure}

\begin{figure*}[tp]
    \centering
    \vspace{0.2cm}
    \setlength{\abovecaptionskip}{-0.1cm}
    \includegraphics[width=17.5cm]{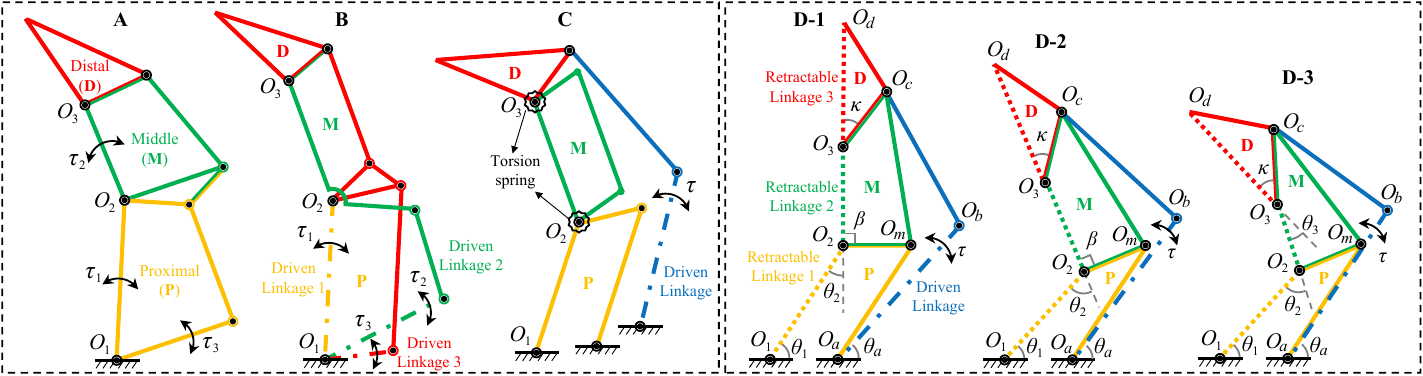}
    \caption{A general three-phalanx finger consists of the metacarpophalangeal (MCP) joint, proximal interphalangeal (PIP) joint and distal interphalangeal (DIP) joint (A); A full-actuated three-phalanx finger (B); An under-actuated three-phalanx finger (C); The prototype of the proposed three-phalanx finger in this work (D). The parallel grasping mode (D-1) and the enveloping grasping mode (D-2/D-3). The thick dotted line represents the driven linkage, and the dotted line indicates that the linkage is retractable.}
    \label{figurelabel}
    \vspace{-0.6cm}
\end{figure*}

Prevalent under-actuated grippers can be broadly categorized as soft and rigid types \cite{grippers_classification,dong4}. The soft grippers are typically constructed via lightweight and flexible materials, driven by tendons or pneumatic/hydraulic systems \cite{Tendon_driven_gunderman,hydraulic_driven_wang,pneumatic_driven}. They prioritize safety, versatility, and adaptability in robotic manipulation, offering a stable grasp of objects with complex contours. For instance, Gunderman \textit{et al.} \cite{Tendon_driven_gunderman} presented a tendon-driven soft robotic gripper with passive compliance, while Wang \textit{et al.} \cite{hydraulic_driven_wang} presented a pneumatic-driven gripper that is able to vary its stiffness and working length. Nevertheless, they have a low load capacity and a short lifespan due to material properties. In contrast, the structure of rigid grippers is dominated by linkage and gear systems, which are known for their stiffness and precision \cite{linkage_driven_kwon,gear_driven}. Toshihiro \textit{et al.} \cite{linkage_driven_Toshihiro} utilized chuck clamping devices to create a gripper. A gripper with fingers that are scissor linkages was proposed by Kwon \textit{et al.} \cite{linkage_driven_kwon}. However, these grippers cannot adjust the grasping range and adapt to objects with irregular shapes, resulting in unstable grasps. To alleviate the above drawbacks, certain grippers were designed by mimicking the human finger to flex and relax muscles for modifying contact surface areas \cite{humanoid_gripper}. This feature significantly improves the graspable range and dexterity of the human hand, allowing it to grasp even slender or thin objects with a single finger. Notably, Bandara \textit{et al.} \cite{humanoid_gripper_Bandara} built an under-actuated gripper based on a modified cross-bar mechanism. Abayasiri \textit{et al.} \cite{humanoid_gripper_Abayasiri} proposed a humanoid robotic hand that employs a 5-linkage mechanism. Yet, these rigid grippers, which are designed to imitate the motion of human fingers, cannot achieve changes in contact surface caused by muscle contraction, which limits their adaptability and envelopment capabilities. To enhance the flexibility and adaptability of robot grippers, some grippers are designed with reconfiguration mechanisms \cite{GTac_gripper,SMS_gripper}. However, most of the reconfigurable grippers rely heavily on additional motors to rotate or translate the fingers, resulting in a complex mechatronic system, such as the GTac gripper \cite{GTac_gripper} and the SMS gripper \cite{SMS_gripper}. Therefore, one current challenge is to develop an under-actuated gripper that has the capability of grasping objects with a large range of objects’ dimensions and high adaptability. Using fewer motors to achieve all the abilities of the gripper is another challenge.

To address the aforementioned issues, we propose a novel under-actuated gripper with three compressible 3-fingers, as shown in Fig. 1. Firstly, the proximal phalanx is composed of a parallel four-bar linkage mechanism to achieve parallel grasping. In particular, each phalanx can realize adaptive retraction through the designed slider-slide rail mechanism to perform enveloping grasping mode, ensuring stable grasping of objects of various volumes and shapes. Secondly, to minimize the number of gripper motors, a transmission mechanism including worm, gears, and racks was artfully designed to distribute the output of the motors. In addition, a slotted block positioned proximal to the finger base is incorporated to constrain the movement of the finger during reconfiguration. Thus, the proposed gripper can achieve both grasping and reconfiguration abilities simultaneously through a single motor. Finally, the presented gripper can actively adjust the grasping range, and passively switch between parallel grasping and enveloping grasping according to the shape of the objects, enabling the gripper to achieve up to five distinct grasping modes. As empirically validated through experimentation, the gripper exhibits notable traits, including elevated rigidity, substantial load-bearing capacity, robust adaptability, and a broad grasping range, which amalgamates the advantageous features of both rigid and soft grippers. 

We \textbf{highlight} the \textbf{novelties} of our work. Foremost, our core contribution to this work is constructing a novel under-actuated gripper with capabilities of high adaptability and a large grasping range of objects’ dimensions. The \textbf{first} novelty is that this work presents an innovative linkage mechanism to achieve the retraction of the phalanges for changing contact areas between the fingers and the objects, thereby varying the friction to improve the grasping stability. Although the proposed gripper is actuated by a single motor, it can perform grasping and reconfiguring simultaneously, considerably reducing the control and mechanism complexities, which is attributed to the \textbf{second} novelty. The \textbf{third} novelty is the capability to achieve five grasping modes by combining finger adaptability and reconfiguration. Different grasping modes can be employed to grasp objects with various sizes and contours, enhancing the stability of the grasp.

\begin{figure}[tp]
    \centering
    \vspace{-0.1cm}
    \setlength{\abovecaptionskip}{-0.1cm}
    \includegraphics[width=8.4cm]{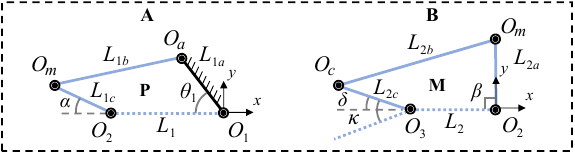}
    \caption{Kinematics model of the proximal phalanx and middle phalanx.}
    \label{figurelabel}
    \vspace{-0.7cm}
\end{figure}
\section{METHODOLOGY}

\subsection{Kinematic Analysis of the Under-actuated Finger}

Three-joint fingers are typically 3 DOF and consist of the MCP joint, PIP joint and DIP joint, they are individually driven by three active linkages, as shown in Fig. 2-A. To facilitate the installation of the motors in fully actuated grippers, the driving linkages are often concentrated at the MCP joint, which is closer to the gripper's palm, as illustrated in Fig. 2-B \cite{figure}. Besides, torsion springs and mechanical blocks can be placed at $\textit{O}_{1}$,$\textit{O}_{2}$ and $\textit{O}_{3}$ to achieve an under-actuated ability, thereby reducing the number of motors (see Fig. 2-C). The proposed gripper utilizes the retractable linkage $\textit{L}_{1}$,$\textit{L}_{2}$ and $\textit{L}_{3}$ to constrain the movement of joints, as depicted in Fig. 2-D. 

Without external contact, retractable linkages $\textit{L}_{1}$ and $\textit{L}_{2}$ are in origin length. In this state, the proximal phalanx ($\textit{O}_{1}$$\textit{O}_{2}$$\textit{O}_{m}$$\textit{O}_{a}$) is maintained in a flexible parallelogram mechanism by $\textit{L}_{1}$, while the middle and distal phalanx combine to form a fixed five-bar mechanism ($\textit{O}_{2}$$\textit{O}_{3}$$\textit{O}_{d}$$\textit{O}_{c}$$\textit{O}_{m}$). Since $\beta$ is a fixed right angle, the orientation of the middle and distal phalanx will remain perpendicular to $\textit{O}_{2}$$\textit{O}_{m}$ as the finger moves. This ensures the parallel grasping mode of the gripper, as shown in Fig. 2-D-1.
\begin{figure*}[tp]
    \vspace{0.2cm}
    \centering
    \setlength{\abovecaptionskip}{-0.0cm}
    \includegraphics[width=17.2cm]{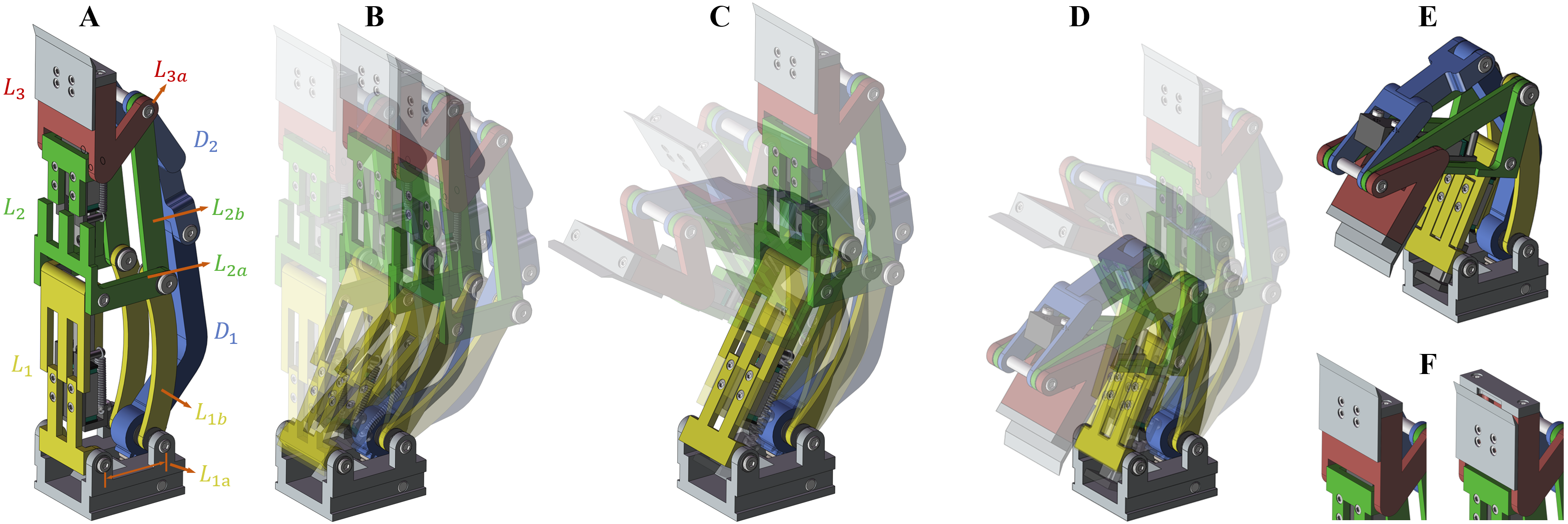}
    \caption{Initial state (A); Parallel grasp (B) and Enveloping grasp (C, D); Maximum flexion state (E); The retractable ability of the distal phalanx (F).}
    \label{figurelabel}
    \vspace{-0.6cm}
\end{figure*}

The angle of $\textit{$\theta$}_{1}$ would be fixed whenever the proximal phalanx makes contact with the objects. The movement of the driven linkage compresses $\textit{L}_{1}$, causing it to change its length.The shape of the proximal phalanx changes, resulting in a shift in the orientation of the middle phalanx.  The middle and distal phalanx remain coupled due to the constraint of the $\textit{L}_{2}$, forming a fixed five-bar mechanism. The gripper switches to the enveloping grasping mode, as indicated in Fig. 2-D-2. 

Similarly, when the middle phalanx comes into contact with the object, the angle of $\textit{$\theta$}_{2}$ is fixed, while $\textit{L}_{2}$ will be compressed. The distal phalanx and middle phalanx will be decoupled. Thus, the distal phalanx can move independently. A finger can bend further to grasp thinner objects, as demonstrated in Fig. 2-D-3. 

Here we construct the mathematical relationship between the angle of each joint and the length of each phalanx after retracting. Taking the proximal phalanx shown in Fig. 3-A as an example, we analyze the length of the retractable linkage $\textit{L}_{1}$ after a finger moves. The vector-loop equation is provided as
\begin{equation}
    \begin{aligned}
        \overrightarrow{\textit{O}_{1}\textit{O}_{a}} + \overrightarrow{\textit{O}_{a}\textit{O}_{m}} + \overrightarrow{\textit{O}_{m}\textit{O}_{2}}= \overrightarrow{\textit{O}_{1}\textit{O}_{2}}
    \end{aligned}
\end{equation}
The phalanx is composed of a four-bar linkage. Except for the retractable linkage $\textit{L}_{1}$, the length of other linkages ($\textit{L}_{1a}$, $\textit{L}_{1b}$, $\textit{L}_{1c}$) is a constant length. And $\alpha$ can be calculated based on the rotation angle of the finger,
\begin{equation}
    \begin{aligned}
        \alpha = \beta - \theta_2 
    \end{aligned}
\end{equation}
A coordinate is built at $\textit{O}_{1}$, with $\textit{O}_{2}$$\textit{O}_{1}$ representing the positive direction of the x-axis. Thus, the coordinates of $\textit{O}_{2}$ ($\textit{o}_{2x}$, $\textit{o}_{2y}$), $\textit{O}_{m}$ ($\textit{o}_{mx}$, $\textit{o}_{my}$) and $\textit{O}_{a}$ ($\textit{o}_{ax}$, $\textit{o}_{ay}$) can be obtained sequentially.
\begin{gather}
    \textit{o}_{2x} = -\textit{L}_{1}, \textit{o}_{2y} = 0 \\
    \textit{o}_{mx} = -\textit{L}_{1} - \textit{L}_{1c}\cos{(\alpha)}, \textit{o}_{my} = \textit{L}_{1c}\sin{(\alpha)}  \\
    \textit{o}_{ax} = - \textit{L}_{1a}\cos{({\theta}_{1})}, \textit{o}_{ay} = \textit{L}_{1a}\sin{({\theta}_{1})}
\end{gather}
To determine $\textit{L}_{1}$, we substitute Eq. (3)-(5) into Eq. (1)
\begin{equation}
	\begin{split}
	\textit{L}_{1b}^2 = (\textit{L}_{1} & +\textit{L}_{1c}\cos{(\alpha)}-\textit{L}_{1a}\cos{({\theta}_{1})})^2\\
        & + (\textit{L}_{1a}\sin{({\theta}_{1})} - \textit{L}_{1c}\sin{(\alpha)})^2
	\end{split}
\end{equation}
After algebraic manipulation, the quadratic polynomial representing $\textit{L}_{1}$ can be derived
\begin{equation}
	\begin{split}
	   \textit{L}_{1}^2 + \textit{b}_{1}\textit{L}_{1}&+ \textit{c}_{1} = 0\\
            \textit{b}_{1}=2\textit{L}_{1c}\cos{(\alpha)}&-2\textit{L}_{1a}\cos{({\theta}_{1})}\\
            \textit{c}_{1} =\textit{L}_{1a}^2+\textit{L}_{1b}^2+\textit{L}_{1c}^2 - &2\textit{L}_{1a}\textit{L}_{1c}\cos{(\alpha-{\theta}_{1})}
	\end{split}
\end{equation}
\begin{figure}[tp]
    \centering
    \vspace{0.2cm}
    \setlength{\abovecaptionskip}{-0.1cm}
    \includegraphics[width=7.3cm]{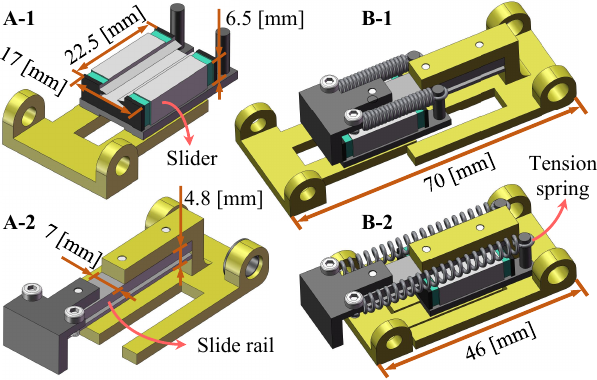}
    \caption{Part connected to the slider (A-1). Part connected to the slide rail (A-2). Retractable linkage at its original length (B-1). Retractable linkage after being compressed (B-2).}
    \label{figurelabel}
    \vspace{-0.4cm}
\end{figure}
Therefore, the retractable linkage $\textit{L}_{1}$ can be calculated by
\begin{equation}
    \begin{aligned}
        \textit{L}_{1} = -\frac 12(\textit{b}_{1}\pm\sqrt{\textit{b}_{1}^2 -4\textit{c}_{1}})
    \end{aligned}
\end{equation}
In Fig. 3-B, the linkage $\textit{L}_{2a}$, $\textit{L}_{2b}$, $\textit{L}_{2c}$ and angle $\kappa$ are constant. Similarly, the length of $\textit{L}_{2}$ can be calculated by
\begin{equation}
	\begin{split}
            \delta = \kappa & - \theta_3 \\            \textit{b}_{2}=2\textit{L}_{2c}\cos{(\delta)} & -2\textit{L}_{2a}\cos{({\beta})}\\
            \textit{c}_{2} =\textit{L}_{2a}^2+\textit{L}_{2b}^2+\textit{L}_{2c}^2 & -   2\textit{L}_{2a}\textit{L}_{2c}\cos{(\delta-\beta)}\\
            \textit{L}_{2} = -\frac 12(\textit{b}_{2}\pm&\sqrt{\textit{b}_{2}^2-4\textit{c}_{2}})
	\end{split}
\end{equation}

The solutions with negative signs are used for determining the lengths of $\textit{L}_{1}$ and $\textit{L}_{2}$.

As previously stated, the inside linkage of each phalanx is the retractable linkage. As shown in Fig. 5-A, each retractable linkage comprises two parts: one part is connected to the MGN7C slider (17×22.5×6.5 mm) and the other part is connected to the MGN7 slide rail (7×4.8 mm). The two parts are constrained by the tension springs. The springs remain unstretched when there is no external force, and the retractable linkage retains its original length. When the phalanx is under stress, the tension springs gradually stretch as the retractable linkage compresses (see Fig. 5-B). To ensure smooth movement of the slider, the springs are installed symmetrically on both sides of the slide rail so that the resultant force is always along the direction of the slide rail and the resultant moment is 0. The types and parameters of the springs at each phalanx are:
\begin{itemize}
    \item MCP joint : tension spring, $\textit{K}_{MCP}$ = 1 N/mm
    \item PIP joint : tension spring, $\textit{K}_{PIP}$ = 0.8 N/mm.
    \item DIP joint : compression spring, $\textit{K}_{DIP}$ = 0.55 N/mm.
\end{itemize}

\subsection{Design of the Under-actuated Finger}
The linkage length parameters for this finger are shown in Table I. Table II illustrates the theoretical parameters for the three retractable linkages after compression. Except for changes in linkage length, the flexion and mutual occlusion of the linkages will further affect the length of the finger contact surface. The actual contact surface length parameters of the phalanges are shown in Table III. It can be found that the maximum compressibility of the phalanges is 57 mm. Which is almost one-third of the original length. The actual contact length can be reduced by 74 mm. The minimum contact length is 57.95\% of the original length.
\begin{table}[h]
\vspace{-0.2cm}
\caption{Linkage parameter}
\begin{center}
\vspace{-0.3cm}
\setlength{\tabcolsep}{4.3mm}{
\begin{tabular}{cc|cc}
\toprule
   Linkage & Length/mm & Linkage & Length/mm \\
   \midrule
    $\textit{D}_{1}$ & 85 & $\textit{L}_{2}$ & 55 \\
    $\textit{D}_{2}$ & 68 & $\textit{L}_{2a}$ & 30 \\
    $\textit{L}_{1}$ & 70 & $\textit{L}_{2b}$ & 76 \\
    $\textit{L}_{1a}$ & 70 & $\textit{L}_{3}$ & 51 \\
    $\textit{L}_{1b}$ & 30 & $\textit{L}_{3a}$ & 29 \\
   \bottomrule
\end{tabular}}
\end{center}
\vspace{-0.6cm}
\end{table}

\begin{table}[h]
\vspace{-0.1cm}
\caption{Analysis of linkage length}
\begin{center}
\vspace{-0.3cm}
\setlength{\tabcolsep}{4.1mm}{
\begin{tabular}{ccccc}
\toprule
   Linkage & $\textit{L}_{o}$/mm & $\textit{L}_{c}$/mm & $\Delta$$\textit{L}$/mm & $\textit{R}_{theory}$\\
   \midrule
    $\textit{L}_{1}$ & 70 & 46 & 24 & 34.29$\%$\\
    $\textit{L}_{2}$ & 55 & 36 & 18 & 32.73$\%$\\
    $\textit{L}_{3}$ & 51 & 36 & 15 & 29.41$\%$\\
    Total & 176 & 119 & 57 & 32.39$\%$\\
   \bottomrule
\end{tabular}}
\end{center}
\vspace{-0.1cm}
\footnotesize{$\textit{L}_{o}$, $\textit{L}_{c}$, $\Delta$$\textit{L}$ represent the original length, the shortest length after compression, and the change in length of each linkage, respectively. $\textit{R}_{theory}$ represents the theory ratio of the change in length.}
\vspace{-0.4cm}
\end{table}

\begin{table}[h]
\vspace{-0.1cm}
\caption{Analysis of contact length}
\begin{center}
\vspace{-0.3cm}
\setlength{\tabcolsep}{3.8mm}{
\begin{tabular}{ccccc}
\toprule
   Phalanx & $\textit{S}_{o}$/mm & $\textit{S}_{c}$/mm & $\Delta$$\textit{S}$/mm & $\textit{R}_{practice}$\\
   \midrule
    Proximal & 70 & 40 & 30 & 42.86$\%$\\
    Middle & 55 & 26 & 29 & 52.73$\%$\\
    Distal & 51 & 36 & 15 & 29.41$\%$\\
    Total & 176 & 102 & 74 & 42.05$\%$\\
   \bottomrule
\end{tabular}}
\end{center}
\vspace{-0.1cm}
\footnotesize{$\textit{S}_{o}$, $\textit{S}_{c}$, $\Delta$$\textit{S}$ represent the original contact length, the shortest contact length after compression, and the change in contact length  of each phalanx, respectively. $\textit{R}_{practice}$ represents the practice ratio of the change in contact length.}
\vspace{-0.3cm}
\end{table}
During the actual grasping process, in the absence of external contact, the finger will keep in parallel grasping mode, as shown in Fig. 4-B. In Fig. 4-C, until the finger contacts the object, the proximal phalanx will retract first. As demonstrated in Fig. 4-D, once the middle phalanx contacts the object, it will start to retract, resulting in the decoupling of the middle phalanx and distal phalanx. Moreover, the distal phalanx also has a retractable ability (see Fig. 4-F), which is serving mainly for grasping thin objects and does not participate in enveloping grasping. When grasping thin objects, this retractable linkage ensures that the fingertips are always in close contact with the object placement surface, making it easier to pick up thin objects.

\begin{figure}[t]
\vspace{0.2cm}
    \centering
    \setlength{\abovecaptionskip}{-0.0cm}
    \includegraphics[width=8.2cm]{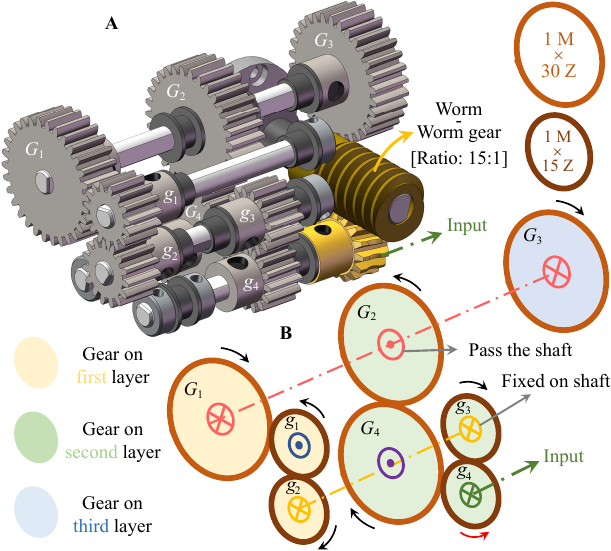}
    \caption{3D modeling(A) and schematic diagram(B) of gear set. }
    \label{figurelabel}
    \vspace{-0.5cm}
\end{figure}

\subsection{Design of the Reconfiguration Mechanism}

The proposed gripper has three fingers arranged crosswise on both sides of the gripper, allowing the gripper to perform grasping and reconfiguration abilities simultaneously with a single motor. The detail of the reconfiguration mechanism is described below:
 
 To drive the fingers distributed on both sides to perform correct motions, a gear set is introduced which is used to allocate the power, as shown in Fig. 6-A. The gear set consists of a set of worm gears with a reduction ratio of 15:1 and several gears with 15 teeth in 1 mold (represented by $\textit{g}$) and 30 teeth in 1 mold (represented by $\textit{G}$). The ultimate goal of the gear set is to achieve the forward rotation of $\textit{G}_{1}$ and $\textit{G}_{3}$ to drive the two fingers on one side, and reverse rotation of $\textit{G}_{2}$ to drive the fingers on the other side. The absolute values of the angular velocities of the three gears are equal, so that the fingers on both sides can flex toward each other at the same speed. 

Following the schematic diagram of the gear set revealed in Fig. 6-B, these gears are arranged in three planes. The gears between different planes are connected by shafts. The same color of the gears implies that the gears are on the same plane. The same color of the gear axis means that the gears are coaxial. The $\otimes$ symbol on the axis represents that the gear is fixed to the shaft and the angular velocity is the same as the shaft, while the $\odot$ symbol on the axis indicates that the gear is only concentric with the shaft and the rotation state is independent of the axis. The motor (proServo-planet PP08D) connected to the worm has a maximum speed of 120rpm and a nominal torque of 6.6Nm. Gear $\textit{g}_4$ is connected to the worm gear, allowing the motor torque to be increased 30 times to 198 Nm. The final motion of each gear is depicted in the figure.

\begin{figure*}[tp]
    \vspace{0.2cm}
    \centering
    \setlength{\abovecaptionskip}{-0.0cm}
    \includegraphics[width=17.5cm]{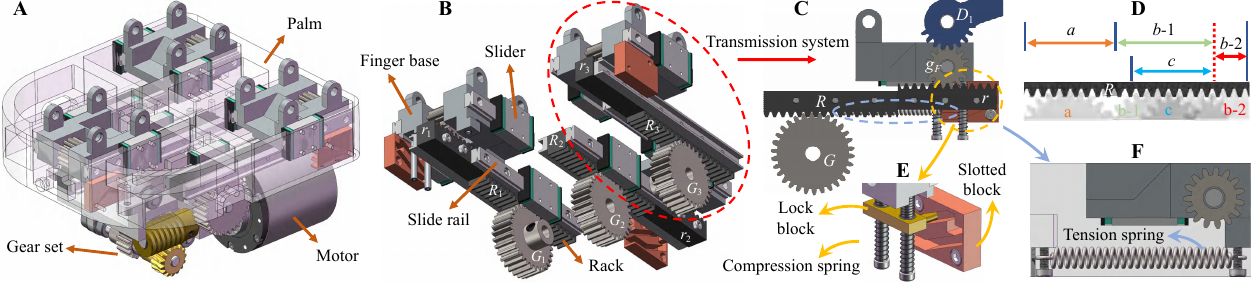}
    \caption{3D modeling of the reconfiguration components (A, B), cross-sectional view of the rack group (C), each part of the rack corresponds to a specific controlled gripping mode (D), self-locking mechanism of reconfiguration components (E) and tension spring of reconfiguration components (F).}
    \label{figurelabel}
    \vspace{-0.5cm}
\end{figure*}

Slide rails are installed on the base of the gripper finger. After assembly on the slide block, the entire translation motion of the finger can be realized, as shown in Fig. 7-A/B. Each transmission system consists of rack $\textit{R}$ (tooth down) and rack $\textit{r}$ (tooth up), they are installed in line and mesh with $\textit{G}$ and $\textit{g}_{F}$, respectively (see Fig. 7-C). $\textit{G}$ represents the three gears ($\textit{G}_{1}$, $\textit{G}_{2}$, $\textit{G}_{3}$) of the final output of the gear set, and $\textit{g}_{F}$ indicates the gear installed in each finger base. The rotation of the motor first drive $\textit{G}$ to translate the rack set. Then, the rack set will drive $\textit{g}_{F}$ to rotate. Finally, $\textit{g}_{F}$ will actuate the $\textit{D}_{1}$ (the driven linkage of each finger), enabling the fingers to move. In addition, a tension spring is installed between the finger base and the palm to constrain the reconfiguration of the finger (see Fig. 7-F). The overall prototype of the proposed gripper is illustrated in Fig. 8. Considering the viewpoint demonstrated in Figure 7-C as a model, we analyze the opening and closing motion of the gripper.

 When $\textit{G}$ rotates clockwise, it will drive the rack $\textit{R}$ and $\textit{r}$ to translate to the right. Due to the constraint of the tension spring, the finger base will initially remain stationary. Thus, $\textit{r}$ will drive $\textit{g}_{F}$ to rotate counterclockwise and linkage $\textit{D}_{1}$ to rotate clockwise. This sequence of actions ultimately leads to the opening of the gripper. Similarly, the counterclockwise motion of $\textit{G}$ will close the gripper. However, when the gripper is fully open, the clockwise rotation of linkage $\textit{D}_{1}$ is limited. Then, $\textit{r}$ will overcome the constraint of the tension spring and drive the entire finger base to move to the right, achieving reconfigurability. It should be noted that after the finger base moves to the far-right end, further counterclockwise motion of $\textit{g}_{F}$ will: 1) move the finger base to the left due to the tension of the spring, 2) actuate the finger performing the closing motion. To achieve the closing action at the far-right end, a self-locking mechanism is introduced for limiting the reverse movement of the finger base, as depicted in Fig. 7-E. This mechanism consists of a lock block, a slotted block and several compression springs. The top and bottom of the lock block are connected to the compression spring. 

\begin{figure}[tp]
    \vspace{0.2cm}
    \centering
    \setlength{\abovecaptionskip}{-0.15cm}
    \includegraphics[width=8.4cm]{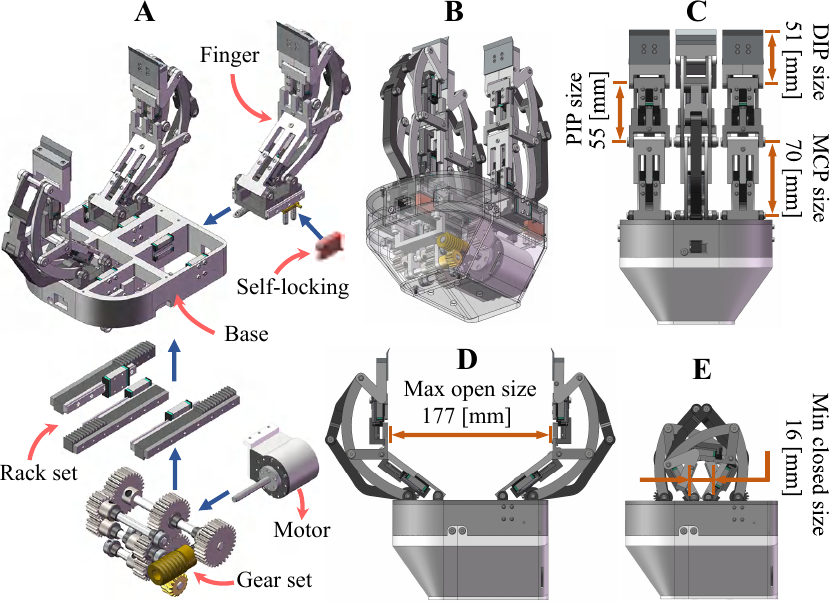}
    \caption{The prototype of the proposed gripper (A), perspective view of driven components (B), the origin length of each phalanx (C), maximum grasping range of parallel grasping mode (D) and minimum grasping range of enveloping grasping mode (E).}
    \label{figurelabel}
    \vspace{-0.50cm}
\end{figure}

\begin{figure}[tp]
\vspace{0.5cm}
    \centering
    \setlength{\abovecaptionskip}{-0.0cm}
    \includegraphics[width=8.4cm]{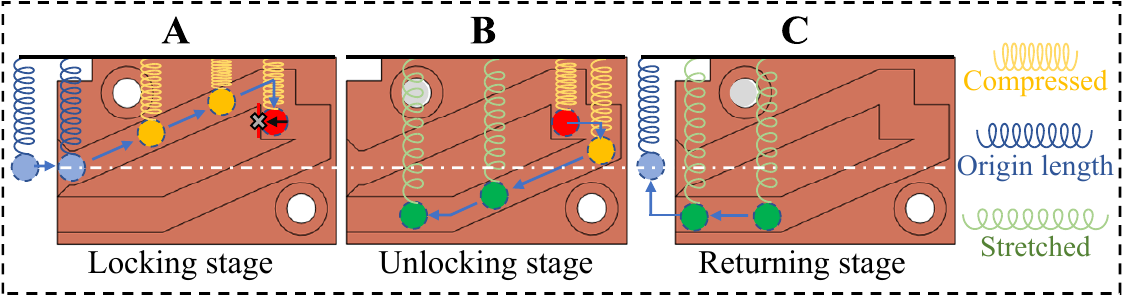}
    \caption{Self-locking flow chart of self-locking mechanism. The circle represents the part of the lock block that extends into the slotted block. The arrow indicates the movement direction of the lock block. The white thin dotted line is the equilibrium position of the lock block. This figure shows the state of the spring on the upper side of the lock block.}
    \label{figurelabel}
    \vspace{-0.6cm}
\end{figure}
 
The specific self-locking process of the mechanism is shown in Fig. 9: 

• Locking stage (Fig. 9-A): The lock block is in an equilibrium position before entering the slotted block. After entering the slotted block, it will move upward along the upper groove, compressing the spring accordingly. When the lock block crosses the top of the incline, the spring pushes it down to the red position. If it changes to move in the opposite direction at this time, it will be restricted by the slotted block and cannot return. Correspondingly, if the finger base stops here, even if $\textit{G}$ rotates counterclockwise, the finger base will not be able to move to the left due to the self-locking mechanism. Therefore, $\textit{r}$ will give priority to driving linkage $\textit{D}_{1}$, realizing the closing motion of the finger at the reconfigured position.

• Unlocking stage (Fig. 9-B): When the lock block is at the red position, continue moving to the right, the lock block will enter the groove below. If the movement is reversed at this moment, the lock block will move along the groove below and will no longer be constrained by the slotted block.

• Returning stage (Fig. 9-C): When the lock block escapes the slotted block, it will return to the equilibrium position with the springs returning to the original length. Rotating $\textit{G}$ counterclockwise will return the finger to its original position.

\begin{figure*}[t]
    \vspace{0.2cm}
    \centering
    \setlength{\abovecaptionskip}{-0.15cm}
    \includegraphics[width=15.5cm]{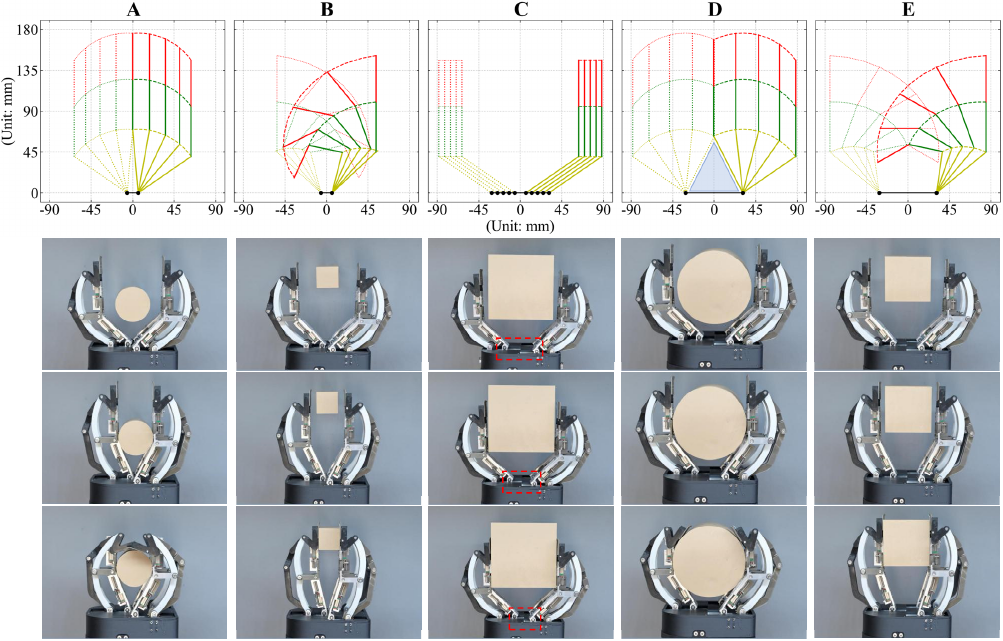}
    \caption{Grasping range and grasping procedure of each grasp mode: proximal parallel grasp mode(A), proximal enveloping grasp mode(B), translational grasp mode(C), remote parallel grasp mode(D) and remote enveloping grasp mode(E).}
    \label{figurelabel}
    \vspace{-0.6cm}
\end{figure*}

\subsection{Description of Multi-grasping Mode}

Combining the described grasping modes of the finger with the reconfiguration  of the finger base, the proposed gripper can achieve a total of five grasping modes. As the gripper is actuated by a single motor, the control strategy is relatively straightforward and does not require an additional motor. Each grasping mode is described as follows:

• Mode 1 (proximal parallel grasp mode): Parallel grasping of fingers in the non-reconfigured state, which is suitable for grasping small regular and thin objects.

• Mode 2 (proximal enveloping grasp mode): Enveloping grasping of the fingers in the non-reconfigured state, which is suitable for grasping slender and small irregular objects.

• Mode 3 (translational grasp mode): The finger uses the translation of the base to achieve grasping. In this mode, the fingertips can only move horizontally, which is useful for grasping thin objects when the fingertips need to be close to the work surface.

• Mode 4 (remote parallel grasp mode): Parallel grasping of the fingers after reconfiguration, which is suitable for grasping large regular objects.

• Mode 5 (remote enveloping grasp mode): Enveloping grasping of the fingers after reconfiguration, which is suitable for grasping large irregular objects.

When $\textit{G}$ meshes with part $\textit{a}$ of the rack in Fig. 7-D, the gripper is in proximal grasping mode. The forward rotation of the motor will control the gripper to open, while the reverse rotation will close it. Notably, the adaptability of fingers enables the gripper to autonomously switch between Mode 1 and 2 based on the shape of the object.

When the gear meshes with part $\textit{b}$-1 of the rack, the motor’s forward rotation will achieve translational grasping mode (Mode 3). 

As soon as the gear $\textit{G}$ meshes with the red dotted line, the self-locking mechanism completes the locking and the gripper completes reconfiguration. On the one hand, the self-locking mechanism unlocks when the motor rotation is reversed, coinciding with gear $\textit{G}$ meshing with the far right end of part $\textit{b}$-2.  The finger will return to its original state from reconfiguration. On the other hand, if reverse the motor immediately after the gear meshes with the red dotted line, the self-locking mechanism will stay locking. Following this, the gear shifts to mesh with part $\textit{c}$ of the rack, prompting the gripper to change to a mode suitable for remote grasping. By keeping the gear meshed with rack part $\textit{c}$, the gripper automatically alternates between Mode 4 and 5 for grasping, depending on the object's shape. Fig. 10 shows the grasping range and the practical application of each grasping mode.

\begin{figure*}[t]
    \vspace{0.2cm}
    \centering
    \setlength{\abovecaptionskip}{-0.1cm}
    \includegraphics[width=17.5cm]{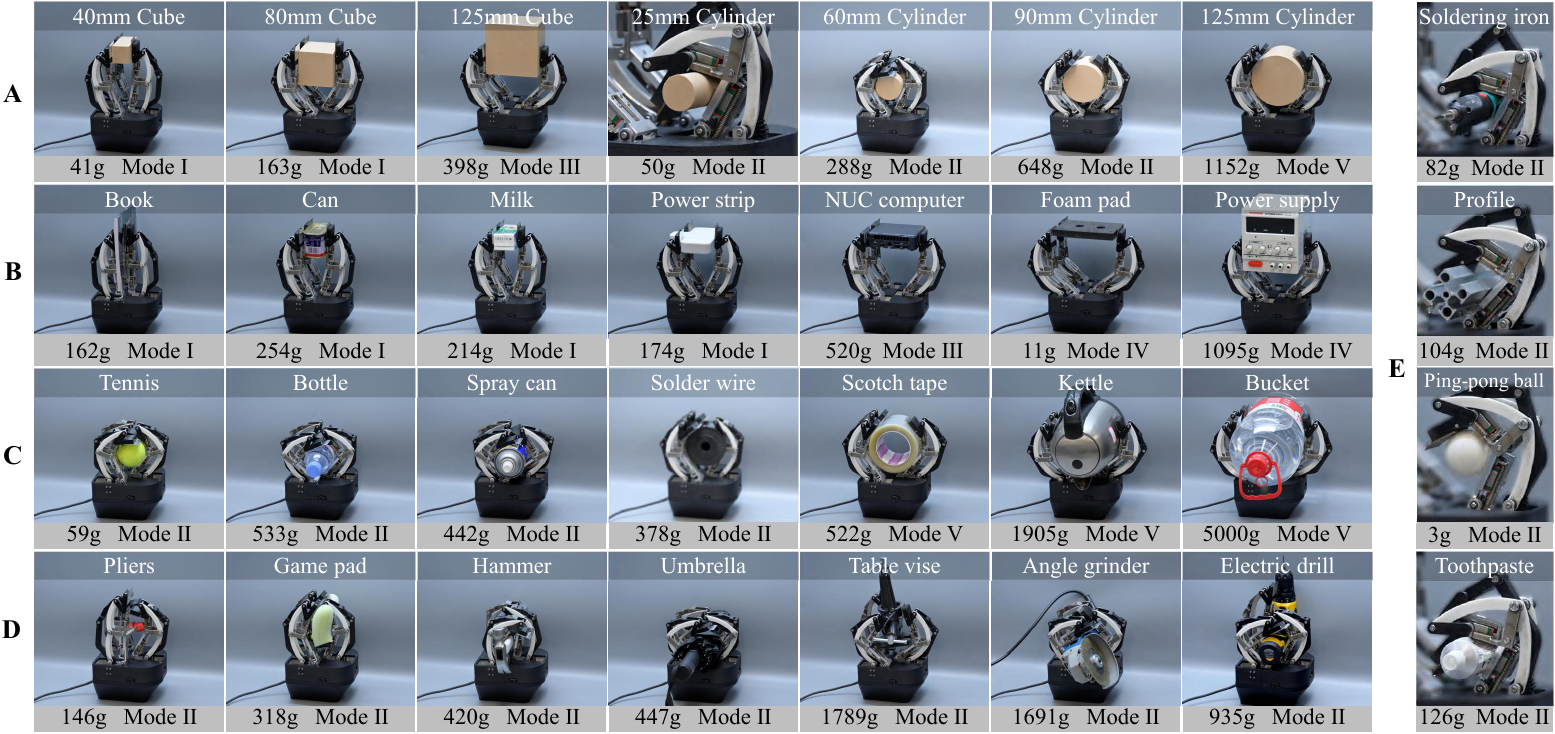}
    \caption{Several demonstrations to verify the grasping performances in different Mode: grasp objects with regular shapes (A); grasp objects with regular planes (B); grasp objects with regular arc surfaces (C); grasp objects with irregular contours (D); grasp slender objects with one finger (E).}
    \label{figurelabel}
    \vspace{-0.5cm}
\end{figure*}

\section{EXPERIMENTS}
To evaluate the grasping performance of the proposed gripper, a prototype three-finger gripper was constructed. A series of experimental validations followed: Firstly, confirmation of the implementation of five grasping modes was conducted. Subsequently, the adaptability of the fingers and the reconfiguration ability of the gripper were verified by grasping regular objects and daily necessities, with the contracted length of each phalanx being measured.

\subsection{Grasps at Different Grasping Modes} 

In the first experiment, the gripper was employed to grasp a variety of objects utilizing five distinct grasping modes. Targets include grasping 60 mm cylinder, 40 mm cube, 125 mm cube, 120 mm cylinder and 80 mm cube in five modes. Detailed grasping procedures are demonstrated in Fig. 10.

During the actual grasping process, it was observed that in the enveloping grasping modes (Mode 1 \& Mode 4), all three joints are involved in grasping objects, resulting in a more stable grasping of the object. Whereas, in the parallel grasping modes (Mode 2 \& Mode 3 \& Mode 5), mainly DIP joints and PIP joints are engaged in grasping objects, leading to precise grasping performance, particularly in the translational grasping mode where the fingertips will not change height in the vertical direction, which enables fixed-point grasping of objects. From another perspective, the switching between parallel grasping mode and enveloping grasping mode (switching between Mode 1 \& Mode 2 and switching between Mode 4 \& Mode 5) is achieved through the retraction of phalanx, which passively adapts the MCP and DIP joints to the contour of the object. In contrast, the reconfiguration mechanism can actively adjust the grasping range by transitioning among the proximal grasping mode, translational grasping mode, and remote grasping mode.

Eventually, the grasping ranges in the five modes are: [0,127]mm, [16,127]mm, [127,177]mm, [0,177]mm, and [34,177]mm. It is noteworthy that although both Mode 1 and Mode 4 fingertips can be completely closed, Mode 4 has a large hollow area after the gripper closes (as marked in the blue area in Fig. 10-D). As a result, grasping small objects may be inhibited in this Mode 4.

\begin{figure}[tp]
    \centering
    \setlength{\abovecaptionskip}{0.2cm}
    \includegraphics[width=8.4cm]{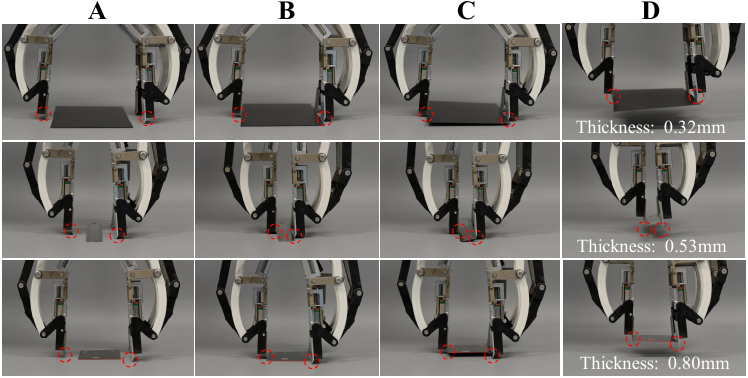}
    \caption{Grasping is achieved with a single finger through the retractable  of the distal phalanx.}
    \label{12}
    \vspace{-0.8cm}
\end{figure}

\subsection{Grasp Objects with Regular Shapes and Daily Necessities}

The proposed gripper was mounted on the 6DOF industrial manipulator-UR3. Subsequent to this, a sequence of grasping experiments was performed to validate the versatility and adaptability of the gripper concerning object manipulation. To evaluate the versatility of the proposed gripper, we applied it to grasp objects of different sizes, shapes, materials, and weights. Fig. 11-A to Fig. 11-D demonstrate the cases of grasping regular objects with regular shapes, regular planes, regular arc surfaces, and irregular contours, respectively. Fig. 11-E illustrates using a single finger to achieve grasping. In this state, the gripper is capable of gripping a 25 mm diameter cylinder. These grasping experiments prove that the gripper has a large grasping range and can adaptively adjust the finger posture according to the contour of the object. 

\begin{table*}[t]
\vspace{0.2cm}
\caption{Analysis of contact area}
\begin{center}
\vspace{-0.3cm}
\renewcommand{\arraystretch}{1.2}
\setlength{\tabcolsep}{4.3mm}{
\begin{tabular}{ccccccccc}
\toprule
   Object & $\textit{L}_{P}$/mm & $\textit{R}_{P}$ & $\textit{L}_{M}$/mm & $\textit{R}_{M}$& $\textit{L}_{D}$/mm & $\textit{R}_{D}$ & $\textit{L}_{total}$/mm & $\textit{R}_{total}$\\
   \midrule
    25mm Cylinder & 47.2 & 32.57 \% & 38.4 & 30.18 \% & 51.0 & 0.00 \% & 136.6 & 22.39 \% \\
    40mm Cylinder & 68.7 & 1.86 \% & 	40.2 & 	26.91 \% & 51.0 & 0.00 \% & 159.9 & 9.15 \%\\
    120mm Cylinder & 68.3 & 2.43 \% & 42.1 & 23.45 \% & 51.0 & 0.00 \% & 161.4 & 8.30 \%\\
   Ping-pong ball & 52.3 & 25.29 \% & 38.2 & 30.55 \% & 51.0 & 0.00 \% & 141.5 & 19.60 \%\\
    Tennis & 60.2 & 14.00 \% &39.4  & 28.36 \% & 51.0 & 0.00 \% & 150.6 & 14.43 \%\\
    Scotch tape & 64.4 & 8.00 \% & 44.8 & 18.55 \% & 51.0 & 0.00 \% & 160.2 & 8.98 \%\\
    Ruler & 69.5 & 0.71 \% & 53.1 & 3.45 \% & 38.2 & 25.10 \% & 160.8 & 8.64 \%\\
    Cardboard & 69.7 & 0.43 \% & 53.6 & 2.55 \% & 39.8 & 21.96 \% & 163.1 & 7.33 \%\\
   \bottomrule
\end{tabular}}
\end{center}
\vspace{-0.1cm}
\footnotesize{$\textit{L}_{P}$, $\textit{L}_{M}$, $\textit{L}_{D}$, and $\textit{L}_{total}$ represent the actual length of the proximal phalanx, middle phalanx, distal phalanx and the total length\, respectively. $\textit{R}_{P}$, $\textit{R}_{M}$, $\textit{R}_{D}$, and $\textit{R}_{total}$ represent the actual ratio of the change in proximal phalanx, middle phalanx, distal phalanx and total length, respectively.}
\vspace{-0.6cm}
\end{table*}

As depicted in Figure 10, the trajectory of the fingertip exhibits an arcuate pattern during parallel grasping or enveloping grasping mode. When grasping an object, the relative height between the fingertip and the object placement surface will change, which can make it hard for the gripper to grasp thin objects, especially since the gesture and position of the gripper are critically required. However, with the retractable  of the phalanx, thin objects can be grasped with ease. As the gripper closes and the fingertips contact the object placement surface, the distal phalanx and middle phalanx adaptively contract to ensure a smooth grasp.  During the grasping process, the fingers are kept close to the surface on which the object is placed, allowing the gripper to better pick up thin objects, as shown in Fig. 12. 

In Fig. 12-A, the fingertips begin to contact the object placement surface. In Fig. 12-B, the gripper adapts to the object placement surface during the closing process. Finally, the picking of thin objects is achieved  (see Fig. 12-C/D). If the paper area is suitable, the gripper can even grasp the paper placed on the desktop.

Table. IV displays the actual length and expansion ratio of each finger phalanx after the gripper grasps objects in enveloping grasping mode. The gripper primarily utilizes the retractable  of the proximal phalanx and middle phalanx to adaptively grasp objects, while using the distal phalanx to grasp thin objects on a desktop. The experiments confirm the feasibility of adaptive grasping by retracting the phalanges of the gripper. The phalanx can shrink up to 32.57\% of its maximum capacity. 

\section{CONCLUSIONS}

In conclusion, this paper investigated a three-finger under-actuated gripper featuring a retractable phalanx based on the linkage system. The proposed gripper not only achieved an extensive grasping range through its reconfiguration ability but also facilitated the automatic switching of grasping modes via finger retraction. Moreover, the implementation of a minimal number of motors objectively mitigated the challenges associated with gripper control. We conducted various grasping experiments to evaluate the real-world performance of the proposed gripper in diverse scenarios. The experimental results demonstrated impressive capabilities in terms of grasping stiffness, versatility, and adaptability. Future work will focus on optimizing the geometric parameters of the gripper, reducing its size and weight, and enhancing the dexterous grasping functionality for smaller objects.

\addtolength{\textheight}{0cm}   


\bibliographystyle{ieeetr}
\bibliography{IROS2024_ref}

\end{document}